\documentclass[runningheads]{llncs}
\usepackage{amsmath}
\usepackage[utf8]{inputenc}
\usepackage[T1]{fontenc}
\usepackage{graphicx}
\usepackage{algorithm}
\usepackage{algpseudocode}

\begin{document}
\title{ZJUNlict Extended Team Description Paper}
\subtitle{\fontsize{9}{10}\selectfont Small Size League of Robocup 2025}
\titlerunning{ZJUNlict Extended Team Description Paper SSL 2025}
%

\author{Zifei Wu \and
Lijie Wang \and
Zhe Yang \and
Shijie Yang \and
Liang Wang \and
Haoran Fu \and
Yinliang Cai \and
Rong Xiong}

\authorrunning{Z. Wu et al.}
%


\institute{State Key Lab. of Industrial Control Technology \\
Zhejiang University \\
Zheda Road No.38, Hangzhou \\
Zhejiang Province, P.R.China \\
\email{rxiong@iipc.zju.edu.cn}
}
\maketitle              
\begin{abstract}
This paper presents the ZJUNlict team's work over the past year, covering both hardware and software advancements. In the hardware domain, the integration of an IMU into the v2023 robot was completed to enhance posture accuracy and angular velocity planning. On the software side, key modules were optimized, including the strategy and CUDA modules, with significant improvements in decision-making efficiency, ball pursuit prediction, and ball possession prediction to adapt to high-tempo game dynamics.

\end{abstract}
\section{Introduction}

ZJUNlict has been actively participating in the competition since 2004, and in 2023, we rejoined the offline world competition after the pandemic, and continued to compete in 2024. Over the past year, we’ve implemented significant hardware and software improvements to enhance our robot’s performance. 

In the hardware section, we focus on the modifications made to the dribbler system, improvements in feedback and prediction, and the integration of IMU control at both the hardware level and the software level. 

In the software section, we present our advancements in ball pursuit prediction, ball possession prediction, the redesign of our strategy module, and optimizations to the CUDA module for real-time decision-making. These developments aim to increase the efficiency, precision, and robustness of our robot’s behavior during high-tempo games, which will be detailed in the following sections.

\section{Hardware}

\subsection{Dribbler Modification}
Dribbling is one of the fundamental abilities of Small Size League robots, determining whether the robot can also move with the ball. Before reaching a stable dribble state, there is an impact at the moment of contact between the dribble ball and the ball. Bouncing occurs when there is a relative velocity between the ball and the robot. Team KIKS [1] confirmed the importance of dampers in improving shock absorption. Team TIGERs [2] designed a 2-DOF structure to improve impact absorption and ball control simultaneously.

\subsubsection{Simplified model}
Initially, we try to drive dynamic differential equations of the dribble system. However, we are confronted with complex non-linear relationships and parameters that are hard to obtain, like the friction coefficient between the ball and the dribble-bar.
To simplify these equations, we abstract the dribble system as a one-dimensional mass-spring-damping system, as shown in Fig.1, where m represents the ball, M represents the dribbler,$k_1$,$c_1$ represent the contact between the ball and the dribbler-bar and $k_2$, $c_2$ represent the damping of the dribbler. In the simplified model, the complex contact between the ball and the dribble-bar is replaced by separable spring and damping.

\begin{figure}[htp]
    \centering
    \includegraphics[width=8cm]{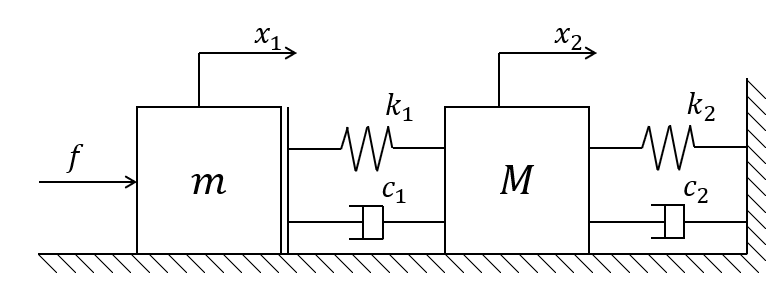}
    \caption{Simplified dribble system model}
    \label{fig:simplified-model}
\end{figure}

Assuming $x_1=0$,$x_2=0$ and $\dot{x}_1=v_0$ at $t=0$ . The ball and the dribble-bar are separated when $x_1-x_2<0$. The dynamic equations are shown as follows.

\begin{equation}
\left\{
\begin{aligned}
  m\ddot{x}_1 &= f \\
  M\ddot{x}_2 &= -k_2x_2-c_2\dot{x}_2
\end{aligned}
\right.
,x_1-x_2<0
\end{equation}

\begin{equation}
\left\{
\begin{aligned}
  m\ddot{x}_1 &= f - k_1(x_1-x_2)-c_1(\dot{x}_1-\dot{x}_2) \\
  M\ddot{x}_2 &= -k_2x_2-c_2\dot{x}_2+k_1(x_1-x_2)+c_1(\dot{x}_1-\dot{x}_2)
\end{aligned}
\right.
,x_1-x_2\geq0
\end{equation}

We can solve these equations numerically in MATLAB and compare the results by analyzing the curve of position of m over time. 

\subsubsection{Simulation and results}
Simulation parameters are shown in Table 1. Since we do not know the true values of each parameter, we can only make qualitative comparisons. To test the influence of M on the simulation results, $M$ = 0.05, $M$ = 0.15 and $M$ = 0.25 are taken respectively. The simulation time step is set to 0.0001s. Simulation results are shown in Fig.2. When M is reduced, the bouncing amplitude of m is reduced and the dribbler system stabilizes faster, which means cutting down the inertance of the dribbler can enhance the dribble ability.Thus, we move the rotational joint to a higher place, as shown in Fig.3.

\begin{table}
\caption{Values of simulation parameters}\label{tab1}
\centering
\begin{tabular}{|l|l|}
\hline
Parameters &  Value\\
\hline
M(kg) & 0.05, 0.15, 0.25\\
m(kg) & 0.046\\
$k_1$(N/m) & 4000\\
$k_2$(N/m) & 100\\
$c_1$(Ns/m) & 5\\
$c_2$(Ns/m) & 20\\
$v_0$(m/s) & 2\\
$f$(N) & 0.13524\\
\hline
\end{tabular}
\end{table}

\begin{figure}[htp]
    \centering
    \includegraphics[width=8cm]{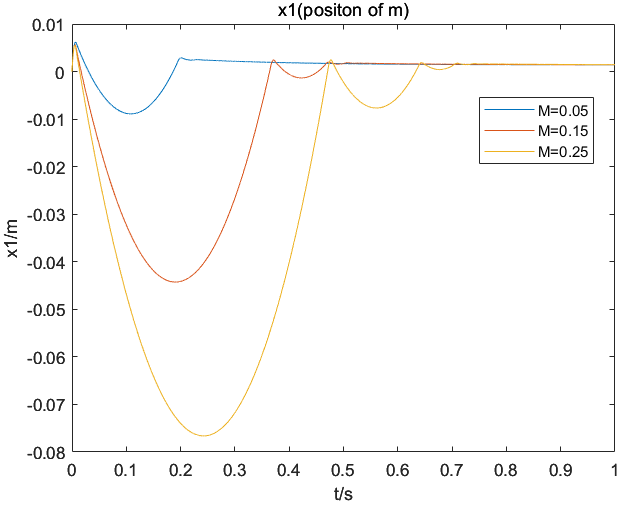}
    \caption{Simulation results of the position of m with different M}
    \label{fig:sim-result}
\end{figure}

\begin{figure}[htp]
    \centering
    \includegraphics[width=8cm]{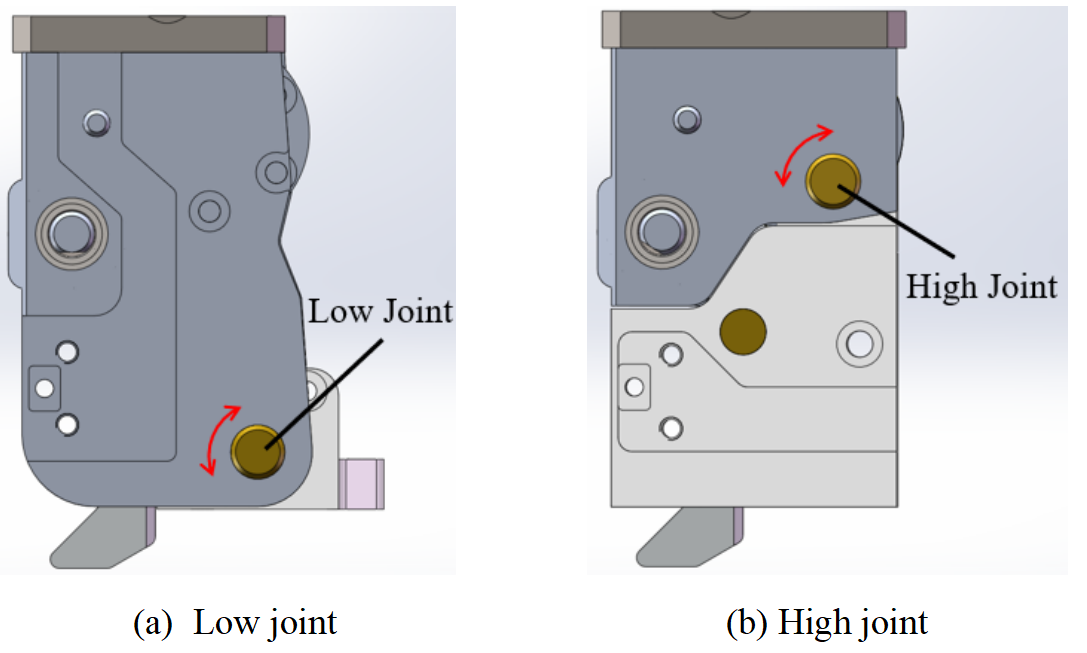}
    \caption{Dribblers of different joint positions}
    \label{fig:joint-comp}
\end{figure}

\subsection{Dribbler Feedback And Prediction}
High-speed dribbling has always been a distinctive feature of ZJUNlict. During competitive matches, high-speed dribbling often enables the robot to intercept the ball and gain control. However, maintaining the stability of the ball spinning at an extremely high speed within the dribbler is not an easy task, as it is significantly influenced by the roughness of the field. The existing infrared feedback from the dribbler is too sparse, and when the ball is close to the robot's dribbler, the vision system cannot determine whether the ball is truly in contact with the dribbler, rendering it ineffective. This makes it difficult to accurately assess the true state of dribbling during matches. Therefore, to address the lack of feedback, we have added an angle sensor to the dribbling drive board to capture the vibration state of the dribbler during dribbling. The angle sensor collects signals at a frequency of 1000 Hz, with 50 signals forming a group. These signals are then processed by an LSTM neural network to predict whether the ball has been successfully dribbled.Considering that the ball dribbling might affect the vibration of the entire robot body, we additionally collected the IMU angle information on the robot to capture potential features for predicting the ball dribbling state. Due to the neural network's reliance on the dataset, we collected separate datasets for dribbling during forward motion, stationary dribbling, rotational dribbling, and backward motion,which is partially shown in Fig.4. With a collection of 5000 datasets, the prediction accuracy on the training set exceeds 90 percent. This prediction result is sent to the software layer as part of the dribbling feedback. 

In earlier implementations, only a single low-cost infrared sensor pair was installed on lateral sides of the dribbler, consisting of one infrared emitter (Model LTE-C9506B) and one infrared receiver (Model PT26-21B/CT). This configuration relied on voltage signal variations to determine ball possession status. However, this arrangement exhibited significant limitations - the sparse single-point feedback could only provide binary ball possession detection (present/absent).To resolve this limitation, this year we deploy 12 infrared sensor pairs linearly along the lateral aspects of the dribbler, arranged in a shallow-to-deep configuration. The combination of infrared sensors at different positions yields high-density feedback signals, which allows us to determine the depth of the ball within the dribbler during dribbling.

\begin{figure}[htp]
    \centering
    \includegraphics[width=10cm]{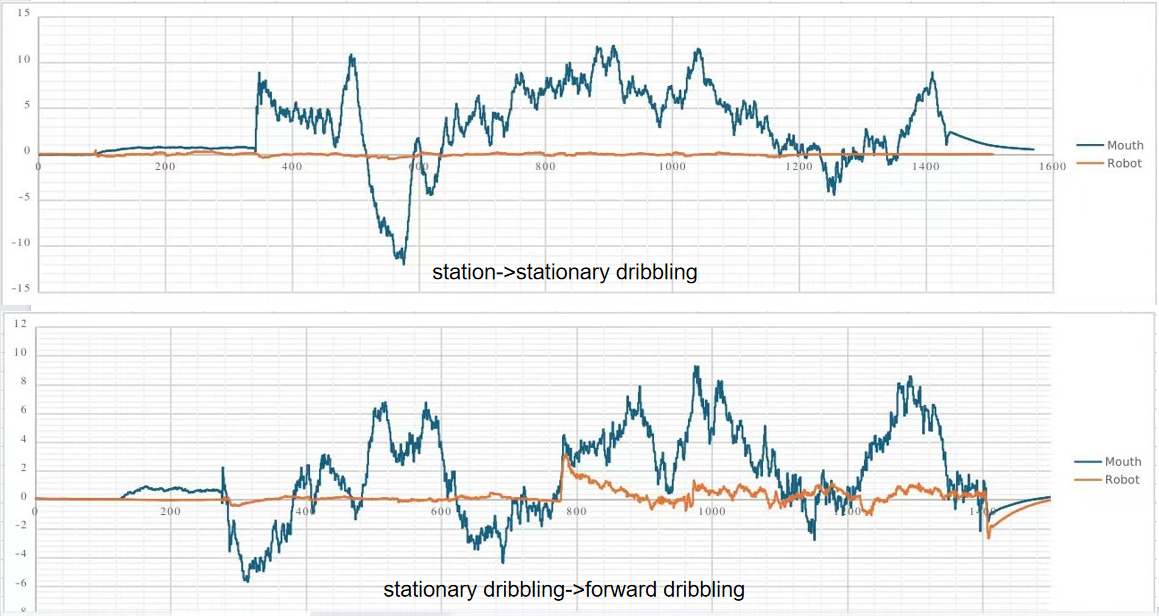}
    \caption{Imu data collected when dribbling}
    \label{fig:dribble_data}
\end{figure}

\section{Software}

\subsection{IMU For v2023 Robot}

Since replacing the robot in 2023[3], we have been utilizing visual information to assess the robot’s posture and plan angular velocities.  This vision-based feedback motion control relies on a 74Hz camera, which severely limits the control frequency and precision within our control architecture. Specifically, our control architecture is hierarchically planned in terms of software and hardware. After the software layer makes decisions and plans, it sends the target speed of each robot to the robot via UDP. This results in the average frequency of visual position feedback being only 74Hz, which is significantly lower than the motor motion control frame rate at the hardware level.  This year, we integrated an IMU into the new robot to obtain posture information with greater accuracy. Once the software layer plans the path and sends the speed to the robot, the robot obtains linear acceleration through the IMU, which is then integrated to derive current linear velocity, enabling high-frequency closed-loop motion control using pid controler, it enables more stable and robust speed planning.

The IMU's coordinate system is initialized when the robot powers on, but it cannot be guaranteed that the IMU coordinates align with the visual coordinates, at the same time, since continuous integration of IMU data to obtain velocity can accumulate errors. Therefore, calibration is necessary to ensure consistency. We correct the IMU coordinates using visual data, ensuring that the IMU coordinate system aligns with the visual system for stable angular velocity planning. In practical, since the IMU is installed horizontally on the robot, the coordinate system calibration is only related to the yaw angle, so only the yaw angle needs to be calibrated. We adjust the IMU's yaw angle based on the visual robot’s yaw angle during each visual update, and no corrections are needed for the angular velocity. The IMU yaw angle is transmitted from the embedded layer to the software layer, enabling the software to access the precise robot orientation. Simultaneously, the target yaw angle is sent to the embedded layer, where the angular velocity is planned.Thus, it is essential to correct the IMU angle in the software layer to align with the visual coordinate system, while also converting the target yaw angle from the visual coordinate system to the IMU coordinate system in the embedded layer. The conversion formula for the IMU yaw angle coordinate system is as follows:
\begin{equation}
\begin{split}
    \Delta \theta = \theta_c^{IMU} - \theta_c^{SSL} \\
    \theta_t^{IMU}= \theta_t^{SSL} + \Delta\theta
\end{split}
\end{equation}

Where $\theta_c^{IMU}$ is the current angle of the yaw angle in the IMU coordinate system, $\theta_c^{SSL}$ is the current angle of the yaw angle in the visual coordinate system, $\theta_t^{IMU}$ is the target angle of the yaw angle in the IMU coordinate system, and $\theta_t^{SSL}$ is the target angle of the yaw angle in the visual coordinate system.

Following the integration of the IMU, we moved angular velocity planning from the software layer to the embedded layer. The advantages of this approach include: 1. Direct access to the accurate angle information provided by the IMU for velocity planning in the embedded layer; 2. Higher planning frequency in the embedded layer compared to the software layer, resulting in reduced state estimation errors and enhanced robustness in robot control.

To evaluate the robot's angular control and velocity planning capabilities, we conducted an experiment to assess the overshoot in motion caused by angular velocity control. The experiment involved moving the robot from a starting point to a fixed straight line, and then along the trackline. The effectiveness of angular velocity control was measured by the maximum offset between the robot's Y-coordinate and the target line. The experimental results, shown in Fig.5, demonstrate that after incorporating the IMU for angular velocity planning, the robot’s Y-direction lateral offset along the trackline was significantly reduced. This indicates that the IMU has improved the robot’s angular velocity planning and angle control, enhancing stability and precision.

\begin{figure}
\centering
\includegraphics[width=10cm]{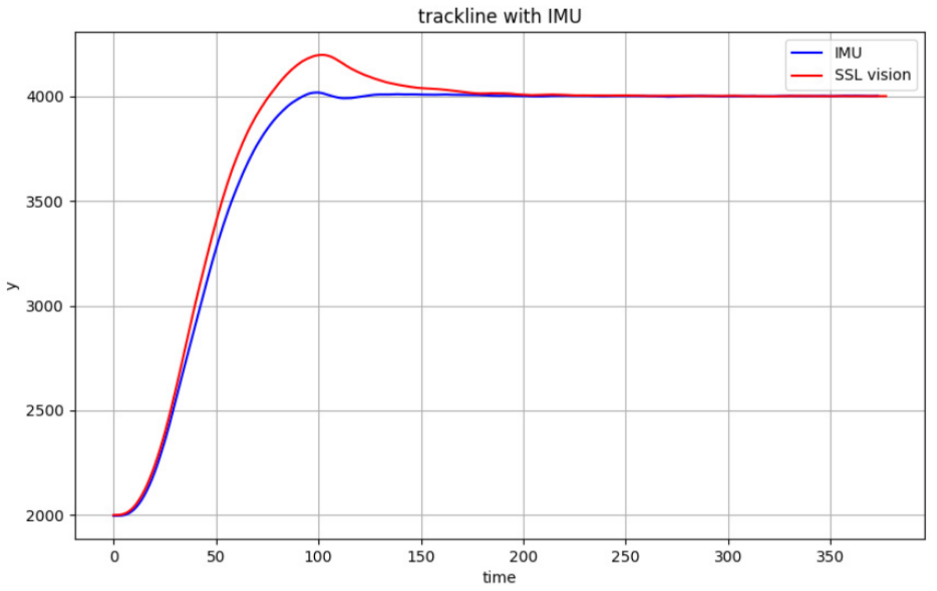}
\caption{Comparison of Y-direction offset along the trackline with angular velocity planning using IMU versus vision-based planning.} \label{fig1}
\end{figure}


\subsection{Ball Pursuit Prediction}

In the past, we developed a series of robot skills for ball pursuit such as self-pass[1], which enhanced our ball possession rate and dribbling efficiency. Based on the search-based interception prediction method from 2019[4], we developed a ball pursuit prediction technique that can forecast the time and final position at which a robot will catch up with and gain control of a moving ball. Considering the stability of the pursuit process, we posit that the relative velocity between the robot and the ball at the moment of contact should be approximately equal. Building on this principle, we designed the algorithm with the following general procedure: first, at each time step, we collect the position, velocity, and other relevant data of both the pursuing robot and the ball. Since the pursuit point must lie on the ball's trajectory, we extend the search starting from the ball's direction. For each point on the trajectory, we calculate the time required for both the robot and the ball to reach that point at the same speed, using our existing ball deceleration model and the robot's trapezoidal motion planning model. These times are denoted as $t_{ball}$ and $t_{robot}$, respectively. Ultimately, when the difference between $t_{ball}$ and $t_{robot}$ for a particular point falls below a predefined threshold, it is deemed that the robot can catch up and control the ball, at which point the algorithm exits the loop.

This algorithm accounts for several special cases that require attention. The first case occurs when the ball has already stopped before the robot can gain control. This situation is represented in the algorithm as the search loop reaching its end, and the stopping position of the ball is directly considered the pursuit point. The second case arises when the ball and robot are moving toward each other. Although the robot's decision-making is more inclined toward intercepting the ball rather than pursuing it in such scenarios, prediction still proceeds. Since the prediction does not inherently account for the need for the robot to maneuver around the ball before pursuing it in the opposite direction, we introduce a maneuvering cost to make the prediction more reasonable. The designed detour cost function takes into account the angle $\theta$ between the robot’s direction and the ball’s velocity vector, as well as the distance $r$ between the robot and the ball. Intuitively, as the robot moves closer to the ball's motion direction and as the distance decreases, the maneuvering cost should increase. Hence, we adopt
\begin{equation}
    \omega_1 e^{-\frac{r}{\omega_2}}(1-\cos\theta)
\end{equation}
as the detour cost function model, where $\omega_1$ and $\omega_2$ are predefined constants. The heatmap in Fig.6 illustrates the results of the predictive algorithm. The prediction scenario is as follows: the ball starts at the coordinate (0, 0) and rolls along the y-axis at speeds of 0.5 m/s and 1.5 m/s, respectively. The color intensity at each point on the heatmap represents the predicted time for the robot to chase the ball to that position (assuming the robot's initial velocity is 0 in the scenario). Lighter colors indicate shorter predicted times, while darker colors correspond to longer times. From the heatmap, it can be observed that the robot requires less time to chase the ball when approaching from behind the direction of the ball's motion. This outcome aligns with both intuitive expectations and actual performance.

\begin{algorithm}
\caption{search-based ball pursuit prediction algorithm}
\begin{algorithmic}[1]
\State \textbf{Require:} $\Delta t$, ball initial position $P_0$ and velocity $v_0$, robot initial position $P_r$ and velocity $v_r$
\State $k \gets 0$
\Repeat
    \State $P_k \gets \text{predictBallPosition}(P_0,V_0,K\Delta t$)
    \State $v_k \gets \text{predictBallVelocity}(V_0,K\Delta t$)
    \State $T_k \gets \text{predictRobotArrivalTime}(P_r,v_r,P_k,v_k$)
    \State $k \gets k+1$
\Until{$|T_k - k\Delta t|\leq t_{thres}$ or $P_k$ out of the field}
\State $P_{best} \gets P_k$
\State $T_{best} \gets T_k$

\end{algorithmic}
\end{algorithm}

\begin{figure}[htp]
    \includegraphics[width=6.4cm]{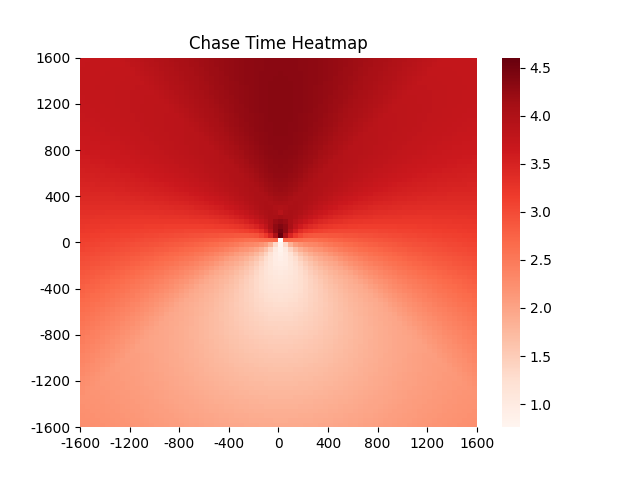}
    \includegraphics[width=6.4cm]{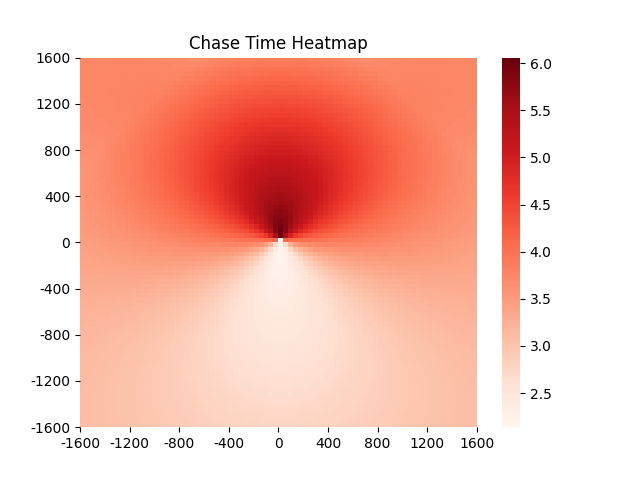}
    \caption{time heatmap with ball speed $0.5m/s$ (left) and $1.5m/s$ (right)}
    \label{fig:heatmap-500-reds}
\end{figure}


\subsection{Ball Possession Prediction}

Ball possession prediction is crucial for our decision-making process, as it determines whether an aggressive or conservative strategy should be employed. In a match, ball possession refers not only to the robot which is currently controlling the ball but also to which team is more likely to gain control of a free ball. We developed a comprehensive ball possession prediction feature that can assess possession status at any given moment on the field. This feature is based on the existing interception and pursuit predictions, with the following general algorithm: initially, we predict both interception and pursuit times for all non-goalie robots on the field. For our team, the decision algorithm uses the same criteria as the actual decision-making process to select either the interception or pursuit time. For the opposing team, the approach is adjusted dynamically based on the specific opponent. Then, we compare the times for the robot with the shortest predicted time to gain possession of the ball for both teams. If our robot’s time is shorter, it is considered our possession; conversely, if the opponent’s robot has a shorter time, possession is attributed to them. Fig. 7 shows the whole process of the possession prediction.

\begin{figure}[htp]
    \centering
    \includegraphics[width=8cm]{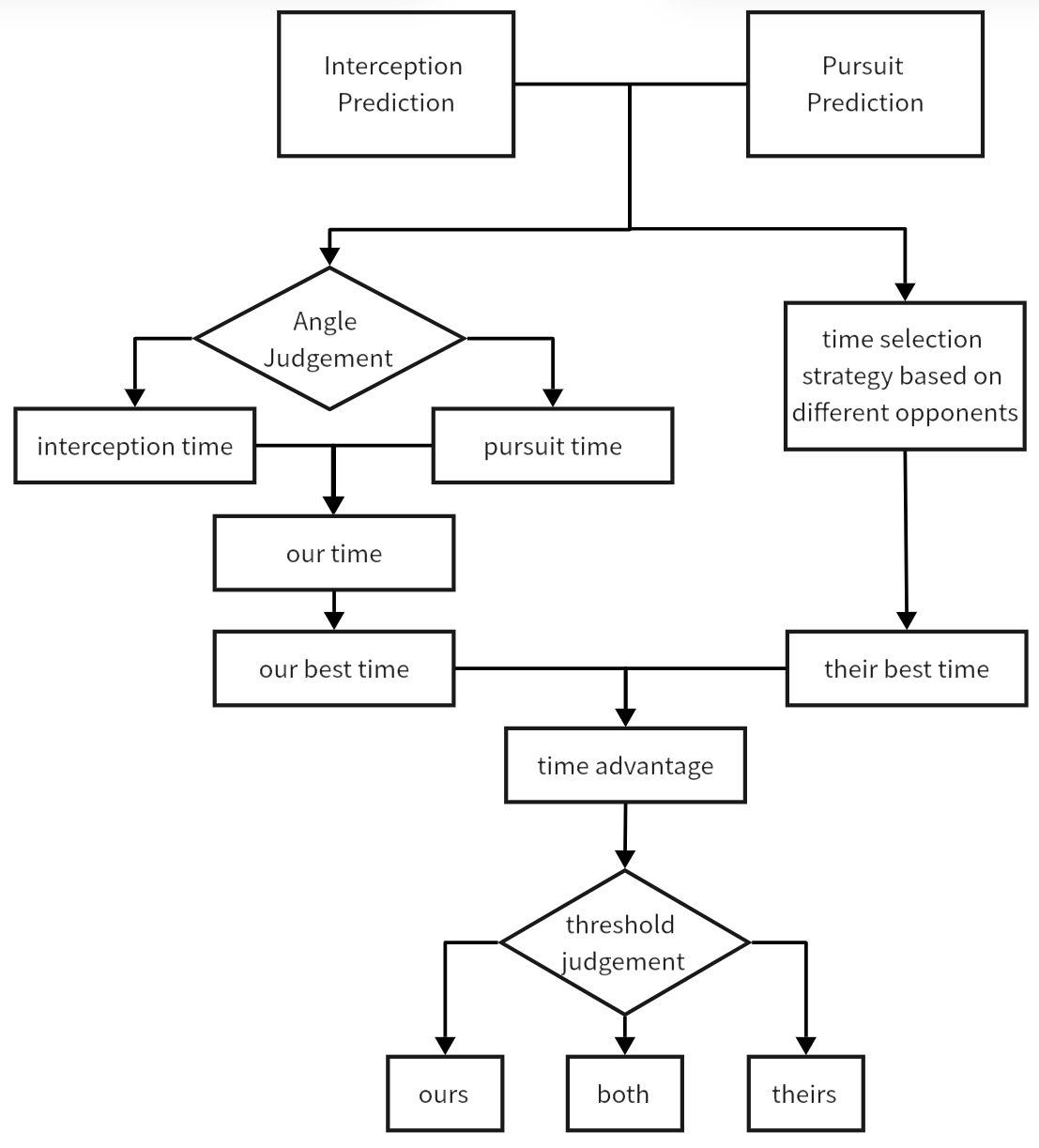}
    \caption{ball possession prediction}
    \label{fig:ball-possession-prediction}
\end{figure}

\subsection{Strategy Module}

In our software framework, the strategy module is located at the top layer and plays a role in overall planning and decision-making. In the past year, we have integrated and optimized our strategy modules, improved decision-making efficiency, and added multiple sets of tactics.

\begin{figure}[htp]
    \centering
    \includegraphics[width=12cm]{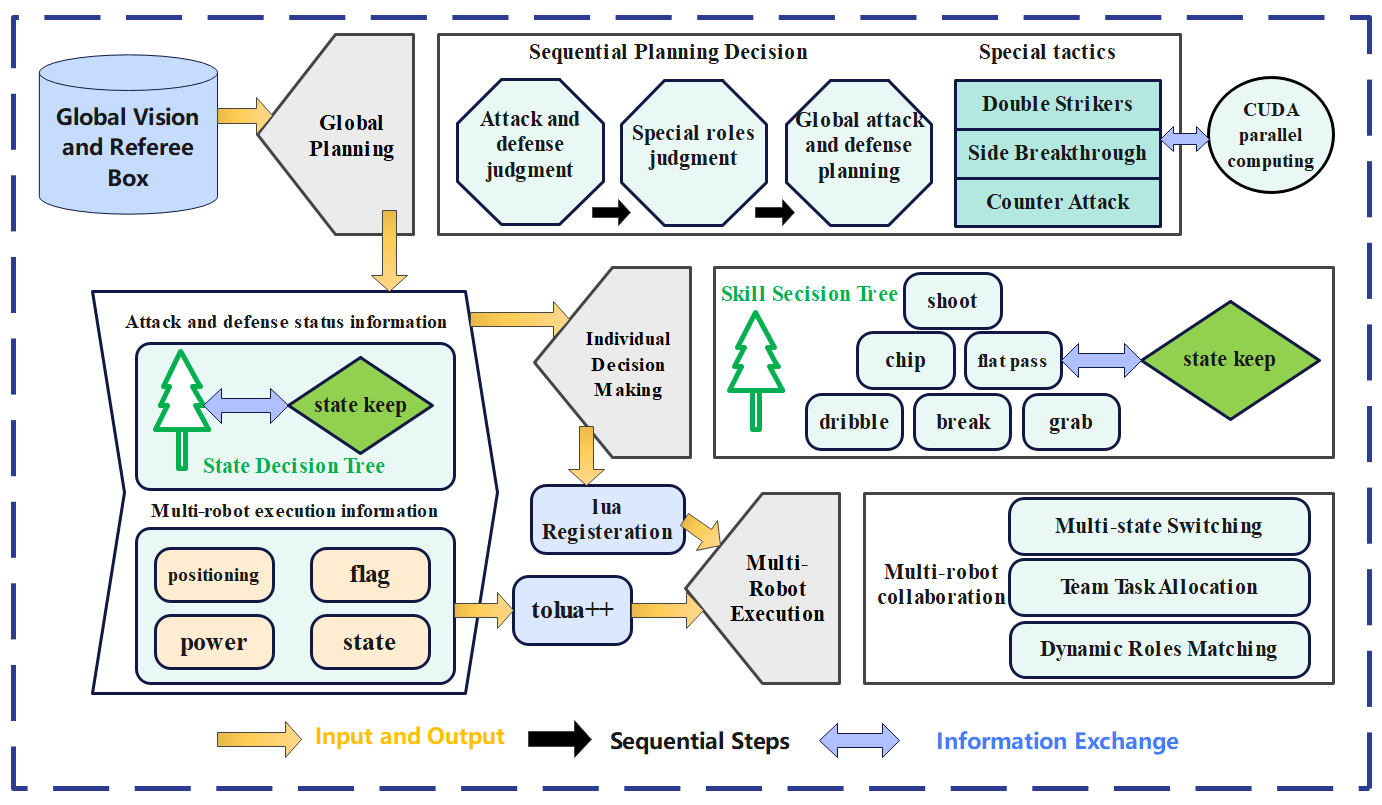}
    \caption{Improved Strategy Module}
    \label{fig:messidecision}
\end{figure}

In the past, our strategy module had a large number of repeated decisions, which were distributed at different decision-making levels. This greatly reduced the efficiency of our decision-making and made the actions that our robot should have insisted on executing inconsistently. To address the issue, we reorganized our strategy module, optimized our hierarchical decision-making framework using the idea of hierarchical control systems, and used the decision tree structure in it to avoid repeated decision-making. We tried to organize the paths between the three levels of our strategy framework so that the three levels only participate in their own decision-making tasks and pass their decision results to the adjacent levels. 

As shown in Fig.8, the top layer, global planning layer takes the global vision and referee box as input, sequentially judges attack and defense state of the match and generates responsive decisions. This layer also considers three special tactics according to the situation on the field to adapt to different opponents. At the same time, since the first layer needs to run at the same frequency of 74 Hz as the visual information, the calculations in it are accelerated by CUDA parallel computing. Individual decision making layer, the second layer, takes the attack and defense decision as input, generates the action decision for a single robot as a skill, the skill includes a continuous sequence of actions, including shooting, chip, flat passe, and dribbling. These actions are organized in the form of decision trees and have state retention mechanism according to the situation. The bottom layer is named as Multi-robot execution layer, it receives the multi-robot execution information of the first level and the skill information registered to lua of the second level, and implements multi-robot execution to play a match through multi-state switching, team task allocation, and dynamic roles matching.

In general, the new strategy module makes decisions more efficient because the paths between levels are clearer and duplicate decisions are avoided. The addition of multiple tactics in the global planning level improves our ability to respond to different game situations.

\subsection{CUDA Module}

CUDA module is an important module in our software module, the passing point, the dribbling point and other important information that describe the situation in match used in our strategy are calculated by CUDA in parallel. Due to frequent and unreasonable calls to CUDA modules in the past, our CUDA module often causes frame rate drops and misses some important passing points, thus we improved the call mode of our software CUDA module.

\begin{figure}[htp]
    \centering
    \includegraphics[width=12cm]{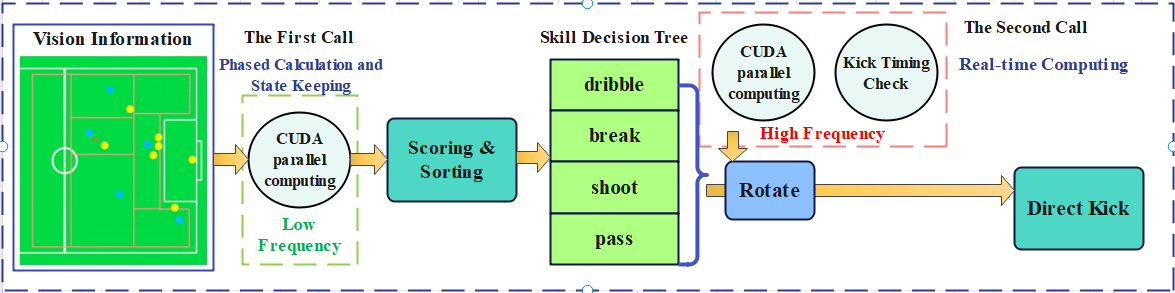}
    \caption{Improved CUDA Module}
    \label{fig:messidecision}
\end{figure}

In the past, the CUDA module was not reasonable enough to calculate the information points calculated by CUDA, such as passing points and passing modes. This is because some regular information does not need to be calculated and updated in real time, such as a passing point in a period. On the contrary, they need to be calculated at a certain frequency and have a certain state retention mechanism to ensure that the robot can persist in performing this task in a short period of time and avoid confusion between the task and the state due to the frequent switching of passing points. For example, our ball-carrying robot should insist within five seconds that it needs to pass the ball to a certain point. The five-second pass point state holding time is used for the robot to complete this action. Different from the regular tasks above, when the ball-carrying car turns around towards the passing point within five seconds, due to the speed generated by the rotation, the ball-carrying car should calculate in real time the landing point of the ball in the current state, and whether teammates can catch the ball at its landing point. These considerations are called direct actions (direct pass and direct shoot) by us, it should be calculated in real time as they are changed significantly every second and therefore we should judge their feasibility with high frequency in our CUDA module.

To address this issue, we changed the CUDA module to a hierarchical two-call mode, as shown in Fig.9. The usual tasks organized into a skill decision tree: dribble, break, shoot, pass are obtained by the first periodic low-frequency CUDA call, accompanied by a state-holding mechanism.

When executing a task in the state decision tree, if the robot is rotating, which means that the angle of the kick and the angular velocity of the kicking ball are constantly changing, we will make a second high-frequency real-time call of CUDA to search for whether the ball can be kicked directly at the current moment in real time, so as to form a direct pass and direct shot, and improve the efficiency of the robot in kicking the ball.


%
%
%
%


\end{document}